# The Benefit of Noise-Injection for Dynamic Gray-Box Model Creation

Mohamed S. Kandil and J. J. McArthur*

Dept. Architectural Science, Toronto Metropolitan University, Canada

*corresponding author: jjmcarthur@torontomu.ca

**Abstract**

Gray-box models offer significant benefit over black-box approaches for equipment emulator development for equipment since their integration of physics provides more confidence in the model outside of the training domain. However, challenges such as model nonlinearity, unmodeled dynamics, and local minima introduce uncertainties into grey-box creation that contemporary approaches have failed to overcome, leading to their under-performance compared with black-box models. This paper seeks to address these uncertainties by injecting noise into the training dataset. This noise injection enriches the dataset and provides a measure of robustness against such uncertainties. A dynamic model for a water-to-water heat exchanger has been used as a demonstration case for this approach and tested using a pair of real devices with live data streaming. Compared to the unprocessed signal data, the application of noise injection resulted in a significant reduction in modeling error (root mean square error), decreasing from 0.68 to 0.27°C. This improvement amounts to a 60% enhancement when assessed on the training set, and improvements of 50% and 45% when validated against the test and validation sets, respectively.

# 1 Introduction

Buildings are significant energy consumers and $CO_2$ emitters and have been consistently identified as a priority for decarbonization by the IPCC [1, 2]. Within buildings, heating, ventilation, and air-conditioning (HVAC) equipment are significant energy consumers and fossil fuel-fired space and water heating systems are the most significant contributors to $CO_2$ emissions. Online optimization of such equipment has demonstrated significant potential for energy savings, energy optimization has demonstrated significant value, with potential energy savings in the range of 25-70% compared with traditional operation [3, 4, 5]. Historically, HVAC equipment simulation to support online optimization has used one of three approaches: white-box models, black-box models, and gray-box models. Each category has its strengths and limitations, making them suitable for different stages and purposes of system analysis. White-box modelling [6], also known as physics-based modelling, relies on a comprehensive understanding of the underlying physical system and consist of mathematical equations that govern the system dynamics. Such models are commonly used during the design phase, where the numerous parameters can be defined. In real-world operation, however, such parameters are unknown and may not exactly match the as-designed condition, resulting in lower accuracy and thus may not suffice for optimization. In contrast, black-box models [7, 8], also referred to as inverse or data-driven models, do not rely on any knowledge of the underlying physics of the system. Instead, they are developed using input and output data collected from measurements. Black-box models are known for their high accuracy and excel in situations where a deep understanding and interpretation of the system's governing dynamics are not necessary. However, they both require a substantial amount of data for training [9] and are limited to predictions within their training domain, leading to significant limitations with regard to the simulation of new controls strategies. Gray-box models offer a compromise between white-box and black-box modelling approaches [7, 8], combining the advantages each by incorporating a physical model structure with unknown parameters. These models utilize measured data from the actual system to estimate these parameters, thereby refining the accuracy of the model. Gray-box models strike a balance between accuracy and data requirements, often needing less training data compared to black-box models. They are particularly useful when a moderate level of understanding and interpretation of the system's dynamics is desired. Additionally, these models can extrapolate beyond the range of training data due to their reliance on physical principles, making them excellent candidates for optimization models.

Despite the inherent benefits of gray-box modeling, they have been shown to possess lower accuracy than black box models when trained on the same data [9]. This paper presents a novel contribution to address this issue, demonstrating the efficacy of deliberate noise-injection to improve dynamic gray-box model accuracy. Unlike previous studies, our research underscores the potential of noise injection to enhance model generalization and address uncertainties common in real physical systems. By introducing controlled randomness through noise injection, our approach empowers the model to explore diverse regions of the parameter space, leading to more robust estimates. We applied this approach in a real-world scenario, using data from a water-to-water heat exchanger serving the heating hot water system of a mixed-use building for training and testing. An identical counterpart served for validation. Our findings, drawn from comparing the model's accuracy trained on unmodified data versus noise-injected data, illuminated the significance of noise injection. Through a series of tests with various noise-injection values, we identified an optimal point consistent across several performance metrics.

This research contributes to the current discourses on both equipment emulation and online equipment controls optimization. We showcase the potential of noise injection in improving parameter estimation and strengthening the reliability and performance of dynamic gray-box models. This innovative approach paves the way for further advancements in parameter estimation methods, ensuring a superior performance of dynamic gray-box models.

This paper is structured as-follows. Section 2 presents an overview of a) modeling of heat exchange equipment, and b) applications of noise injection. Section 3 presents the methodology, including a description of the test apparatus, the governing mathematical equations for the equipment, and the approaches used for parameter estimation, noise injection, and performance evaluation. Results are discussed in Section 4, comparing the impact of noise injection on heat exchanger prediction accuracy, followed by a broader discussion of the implications of this research in Section 5 and concluding summary, along with insights on future work in Section 6.

## 2   Literature Review

To contextualize this study, a review of two topics of core relevance to this paper is presented in this section. Initially, we explore various HVAC modeling techniques, followed by a discussion on noise injection. These are presented in the following sections to better locate this study within the literature and provide insight both on the existent research gap addressed by this paper and to justify the methodology presented.

### 2.1   Modeling Heat Exchange Equipment

Several review papers provide an excellent overview of the HVAC system modeling domain. A comprehensive review in [7] focused on data-driven techniques including gray-box and black-box models. They discussed the respective advantages and limitations of each category in the context of studying energy consumption and control strategies. Similarly, a comprehensive review was conducted in [8] to assess the suitability and performance of various modeling techniques used in

HVAC systems for improving energy efficiency and indoor comfort through advanced control strategies. The review identified strengths, weaknesses, and real-world applications of these techniques, along with recommendations to enhance HVAC system performance. A comprehensive review of black-box and gray-box modeling approaches for HVAC systems, focusing on AHU and indoor thermal load was presented in [10].

The modeling of heat exchangers is typically categorized into three primary approaches: white-box, black-box, and gray-box models [8]. Moreover, they can be further subdivided into steady-state and dynamic models. Dynamic models find extensive use in advanced control strategies and fault detection algorithms, contributing to the enhancement of HVAC system performance and its fault detection and diagnostic (FDD) capabilities [11]. In contrast, steady-state models offer simplicity and computational ease. Their straightforward nature makes them particularly suitable for basic applications where rapid calculations and a general system overview are essential. However, it's worth noting that while steady-state models offer computational advantages, they may not capture the intricate temporal variations observed in dynamic models.

Heat exchangers can be grouped into three primary types: air-to-air, air-to-water, and water-to-water. The air-to-air type, often found in air handling units (AHUs), transfers heat between air streams, ensuring conditioned air in buildings with minimal cross-contamination. Notably, energy recovery ventilators (ERVs) under this category boost energy efficiency by reclaiming both heat and moisture from exhaust air. Air-to-water variants facilitate heat exchange between air and water, predominantly aiding space heating and cooling in HVAC systems. Meanwhile, water-to-water types, frequently seen in domestic hot water systems, hydronic heating, and various industrial processes, involve heat transfer between two water streams. The common designs for this type are shell and tube and plate heat exchangers. Regardless of their distinct purposes and forms, these heat exchangers can generally be modeled using energy balance equations [7, 8]. A simplified ordinary differential equation model applicable to all heat exchanger types (air-to-water, air-to-air, and water-to-water) was used in [12]. For consistency with the heat exchanger type examined in this study, a water-to-water model is outlined as follows:

$$Cz \cdot \frac{dT_{co}}{dt} = m_h \cdot c_w \cdot (T_{hi} - T_{co}) - UA \cdot \left( \frac{T_{co} + T_{hi}}{2} - \frac{T_{ho} + T_{ci}}{2} \right) \quad (1)$$

$$Cz \cdot \frac{dT_{ho}}{dt} = -m_c \cdot c_w \cdot (T_{ho} - T_{ci}) + UA \cdot \left( \frac{T_{co} + T_{hi}}{2} - \frac{T_{ho} + T_{ci}}{2} \right) \quad (2)$$

where $C_z$ is the thermal capacitance of the heat exchanger (units: J/°C). $T_{co}$ is the temperature of the water leaving the heat exchanger at the cold side (units: °C). $T_{hi}$ is the temperature of the water entering the heat exchanger at the hot side (units: °C). $T_{ho}$ is the temperature of the water leaving the heat exchanger at the hot side (units: °C). $T_{ci}$ is the temperature of the water entering the heat exchanger returning at the cold side (units: °C). $m_h$ is the mass flow rate of the water at the hot side (units: kg/s). $m_c$ is the mass flow rate of the water at the cold side (units: kg/s) $c_w$ is the specific heat capacity of water (units: J/(kg·°C)). $UA$ is the overall heat transfer coefficient times the heat transfer area (units: W/°C).

White-box modeling for HVAC equipment is typically conducted using building performance tools such as TRNSYS [13] or broader simulation platforms like Modelica [14]. While these platforms have rich HVAC component libraries [15], they often fall short in the flexibility for controller development, especially for advanced systems like model predictive control (MPC). Transitioning to versatile programming environments like Matlab Simulink, Python, or Julia can better support the creation and integration of advanced controllers and energy-saving strategies

[16, 17]. Parameters for white-box models can be derived from manufacturers' catalog data [18, 19] or through on-site tests [20]. Nevertheless, white-box models come with their own set of challenges. These include inherent complexity, the need for an in-depth understanding of the system, and often a lack of manufacturer-specific data [8].

Black-box models encompass a variety of approaches, including frequency domain models with dead time, data mining algorithms, fuzzy logic, statistical models, state-space models, case-based reasoning models, geometric models, and instantaneous models [7]. In [21], the authors utilized step response-based system identification to derive the transfer function for a laboratory heat exchanger, which was subsequently used for the development of a robust MPC. The integration of neural networks with MPC to capture complex system dynamics and achieve efficient energy savings was proposed in [22]. Additional data-driven models for control purposes can be found in references [23, 24, 25, 26]. Data-driven models have been extensively utilized for fault detection and diagnosis (FDD) [27, 28, 29].

Gray box models blend the strengths of both physics-based and data-driven models. They employ physics-based methods to structure the model while using system performance data to estimate parameters. Afram and Janabi-Sharifi [9,12,30,31] undertook a comprehensive series of research investigations centered on the modeling of residential HVAC systems, particularly those involving heat exchange-based equipment. In one study [12], they developed a gray-box model that combined first principles and empirical data. They employed nonlinear least squares optimization to determine the model parameters. While this model accurately captured the dynamics of a specific HVAC system under certain conditions, its validation was confined to a training dataset, with no assessment on test or validation sets. The paper also lacked clear criteria for the selection of initial conditions, which were essential for optimization convergence. In another study [9], the researchers investigated various black-box modeling approaches commonly used in Canadian residential HVAC systems. These included artificial neural networks (ANN), transfer function models (TF), process models, state-space models, and auto-regressive exogenous (ARX) models. Compared to their previous gray-box models, these black-box models consistently outperformed in terms of both training and validation datasets. Similar conclusions were drawn in a follow-up paper [30] that directly compared the performance of gray-box and black-box models across different seasons. In a more recent paper [31], Afram and Janabi-Sharifi refined their gray-box modeling technique by addressing previous limitations. They implemented a pattern search optimization using the Latin hypercube method and the Active-Set algorithm, thereby eliminating the need for manual selection of initial conditions. This led to two distinct parameter sets optimized for winter and summer conditions. Performance metrics, including coefficients of correlation, determination, and goodness of fit, showed a high level of precision, with values exceeding 70% and reaching up to 88% in some instances. Notably, however, the paper did not include a comparative analysis with their earlier optimization techniques based on nonlinear least squares algorithms. Further gray-box models can be found in [32, 33].

In this subsection, we provided an overview of the literature concerning the modeling of heat exchange equipment, with a specific focus on the dynamic gray-box models due to its pivotal role in constructing models for advanced control and FDD applications. Recognizing its significance, we put forth the concept of employing noise-injection as a solution to boost the accuracy of dynamic models. The subsequent subsection presents the noise-injection technique, along with its practical implementations across the realms of control and machine learning.

## 2.2 Applications of Noise-Injection

The noise-injection technique [34], sometimes referred to as "stochastic perturbation", involves the deliberate addition of random variations, or noise, to the data being processed by the model. This practice has its roots in the field of stochastic differential equations where noise is an inherent part of the model, representing natural randomness in the system being modeled [35].

In the realm of machine learning, noise injection finds widespread application, particularly in training neural networks, to enhance model performance and robustness [36, 37, 38]. By introducing small, random variations to the data during processing, the model becomes more resilient and can generalize better to unseen data. Various noise injection techniques are employed in neural networks, including input noise, weights noise, activation noise, or gradient noise [39, 40, 41]. This practice creates a more challenging environment for the model during training, not unlike teaching the model to handle worst-case scenarios. This can prevent overfitting to the exact data and can help the model learn to handle minor fluctuations or variations in the data, leading to a more robust and generalizable model and improved real-world performance with imperfect data.

Deep reinforcement learning (RL) methods often explore through noise injection in the action space [42]. However, an alternative approach involves adding noise directly to the agent's parameters. This method can result in more consistent exploration and a wider range of behaviors, making it a promising option for enhancing RL performance.

Noise-injection is also used in control applications [43, 44] including adaptive control, system identification, online parameter estimation, and Model Predictive Control (MPC) [45]. This process, referred to as "input/output excitation" or "system excitation" involves adding noise to the system inputs or perturbing the outputs. The purpose is to gather diverse data, improve model accuracy, and enhance control performance while ensuring robustness in the face of uncertainties.

The injected noise should possess persistence of excitation to effectively explore the system's behavior and make better-informed decisions. To date, there is a paucity of literature exploring such noise injection into HVAC equipment models.

## 2.3 Summary of Literature Review

In the realm of HVAC modeling, researchers have often delved into steady state and dynamic models, employing a variety of techniques ranging from white box to gray box and black box models. By integrating the best of white-box and black-box techniques, gray box models achieve superior generalization compared to data-driven models and enhanced accuracy relative to physics-based models. However, they are fraught with challenges like model nonlinearity, unmodeled dynamics, and susceptibility to local minima which could result in poor generalization [9]. In an attempt to mitigate these issues, our research introduces controlled randomness into the model through noise injection. This technique improves the model's ability to explore a broader parameter space, thereby providing more robust estimates. The idea behind this is akin to techniques like data augmentation in machine learning: by introducing variations into the data (in a controlled manner), a model can become better generalized and more robust to real-world uncertainties and disturbances.

Choosing IPOPT (Interior Point OPTimizer) [46] as our solver brings its own set of implications. IPOPT's strengths lie in handling large-scale, continuous, nonlinear optimization problems, making it apt for our project. Yet, the introduction of noise could both aid and complicate the solver's performance. It may offer a form of regularization, guiding IPOPT toward more generalized solutions, or create new local minima, complicating the solver's job. Additionally, while noise can make certain landscapes more tractable for IPOPT, excessive or poorly chosen noise can also result in the model losing its generalization capabilities. Hence, iterative testing and post-training evaluation on clean data are indispensable steps in the methodology. Accessibility to high-level computational resources is another consideration in our research. Thanks to the NEOS server [47], a platform that democratizes access to a multitude of powerful solvers, researchers can efficiently engage with optimization problems without the complexities of local installations and operating system dependencies.

In subsequent sections, we delve into several critical considerations surrounding the use of noise injection in dynamic gray-box models. First and foremost, the noise incorporated should closely reflect the uncertainties and disturbances that the system is likely to encounter in real-world scenarios. Injecting random noise indiscriminately could be counterproductive or even harmful to the model's performance. Additionally, while noise can enhance parameter estimation, it is not a remedy for fundamental flaws in the model's structure. If the model lacks the complexity to capture the system's inherent dynamics, the introduction of noise won't rectify this limitation. When it comes to optimization problems, noise can sometimes improve tractability, particularly if the solver has difficulty navigating specific types of landscapes, such as flat regions. However, this comes with its own set of challenges, such as the introduction of new local minima where the solver could become trapped. Finally, it's imperative to validate the model using clean, real-world data post-training. This ensures that any improvements in model performance are genuine and not merely an artifact of the injected noise. The validation of our methodology is supported by the promising results obtained from developing models for water-to-water heat exchangers and benchmarking them against conventional techniques, demonstrating the potential to improve parameter estimation techniques and ensure the reliability and effectiveness of dynamic gray-box models. This innovative method opens new avenues for advancing parameter estimation techniques and elevates the overall performance of dynamic gray-box models.

## 3 Methodology

To develop this approach, we consider a water-to-water plate heat exchanger system for a real building heating system, described in Section 3.1, and create a mathematical model for its operation (Section 3.2), drawing from the literature. Data was then collected from real system and processed as described in Section 3.3. Parameters were estimated following the procedure outlined in Section 3.4, both without noise injection, and again with noise injected as described in Section 3.5. Finally, Section 3.6 discusses how the performance was evaluated for each model.

### 3.1 System Description

The hydronic heating system at The Daphne Cockwell Complex (DCC) in Toronto, Ontario, is a remarkable dual-purpose building designed for education and residence. It houses academic spaces, administrative offices, student accommodations, and other facilities, creating an environment that fosters collaboration, innovation, and interdisciplinary work. The building's unique design integrates living and learning spaces, with the first 8 floors dedicated to academic purposes and the remaining 19 floors accommodating residence apartments, offering a holistic approach to education and student life.

The heating system serving the academic podium comprises two plate heat exchangers, installed in parallel, which provide high-temperature hot water to the building (the demand or 'cold' side) from a central heating water loop (the supply or 'hot' side). These heat exchangers are extremely common as they facilitate the transfer of heat from one water stream to another) without any fluid mixing and without any internal moving parts. In this setup, hot water from the heat source, in this case the district energy system, forms the supply loop and is connected to the hot side inlet of the heat exchanger. It flows through the heat exchanger, transferring heat via the metal plates to the demand loop, which circulates heating water to the building. This configuration optimizes the building heating hot water supply temperature, maintaining it at the desired 60°C level for perimeter heating. Figure 1 shows the schematic for this setup, where $m_{h\_tot}$ and $m_{c\_tot}$ are the total hot side and cold side water mass flow rates, respectively, $T_{hi\_x}$ and $T_{ho\_x}$ are the hot (supply) side inlets and outlets on each heat exchanger (X={1,2}, and $T_{ci\_x}$ and $T_{co\_x}$ are the cold (demand) side inlets and outlets on each heat exchanger.

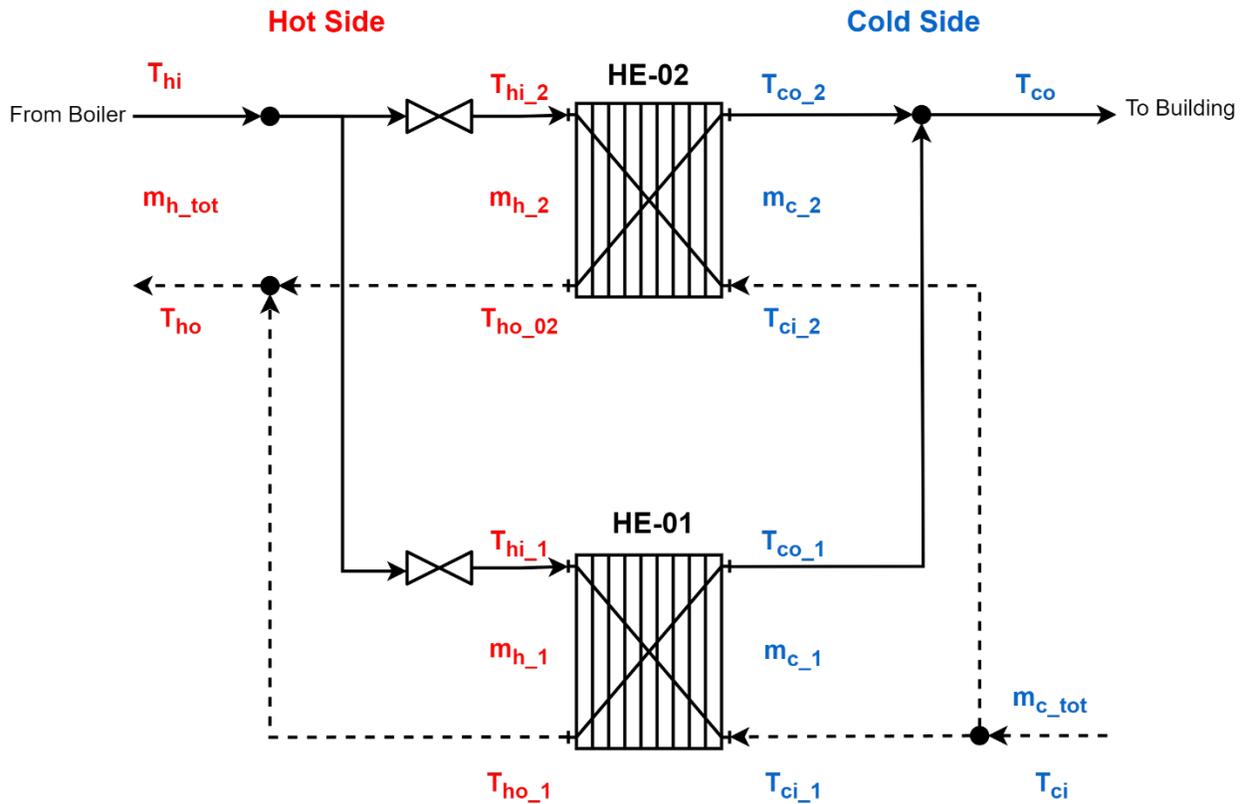

*Figure 1 Control schematic with simplified annotations for the heat exchanger system modeled in this study*

## 3.2 Mathematical Modeling

The heat exchanger can be effectively modeled by establishing an energy balance equation for both the supply and return water streams. Ordinary Differential Equations (ODEs) are employed to represent the dynamics of the heat exchanger (HE) in the form of a lumped parametric model.

This model captures the essential behavior of the HE system and simplifies its representation for analysis and simulation purposes. It is worth noting that since the two HEs are identical, we need only provide one set of equations below to represent a single HE. For simplicity, we will omit the subscript numbers that reference the HE. We also assume that $m_h = 0.5 \cdot m_{h\_tot}$ and $m_c = 0.5 \cdot m_{c\_tot}$. The inputs and outputs of the gray-box model is illustrated in Figure 2.

To model the heat transfer and simplify the parameter estimation process, we can rewrite the equations (1) and (2) as follows:

$$\frac{dT_{co}}{dt} = k_1 \cdot m_h \cdot (T_{hi} - T_{co}) - k_2 \cdot \left(\frac{T_{co} + T_{hi}}{2} - \frac{T_{ho} + T_{ci}}{2}\right) + k_3 \tag{3}$$

$$\frac{dT_{ho}}{dt} = -k_1 \cdot m_c \cdot (T_{ho} - T_{ci}) + k_2 \cdot \left(\frac{T_{co} + T_{hi}}{2} - \frac{T_{ho} + T_{ci}}{2}\right) + k_4 \tag{4}$$

where $k_1 = c_w/C_z$ and $k_2 = UA/C_z$. The parameters $k_3$ and $k_4$ have been added to each equation to account for any unmodeled dynamics or to calibrate sensor offsets. Therefore, the parameters that need to be identified are $k_1$ to $k_4$. The inputs and outputs to the model can be represented with vectors $u = [T_{hi}, T_{ci}, m_h, m_c]$ and $x = [T_{co}, T_{ho}]$ respectively.

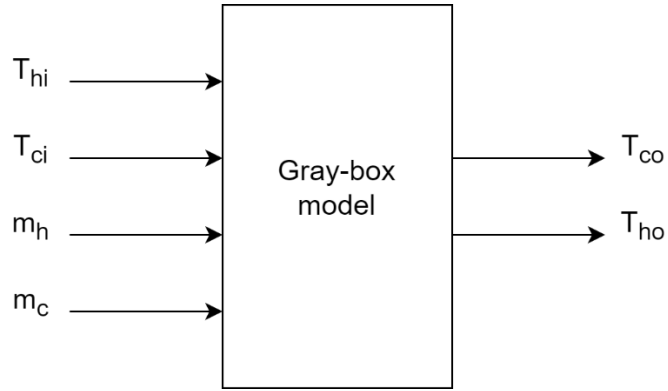

*Figure 2 Block diagram illustrating the inputs and outputs of the gray-box model*

### 3.3 Data Collection and Pre-processing

In this study, data were collected from the case study system in December 2022, representing normal post-lockdown operation of the facility. The recorded data was stored in cloud-managed time-series database (AWS Timestream DB). The data stream was configured to output each data point at each change of value (threshold of twice the sensor precision), resulting in an irregular sampling rate. To tackle this irregular sampling issue and enhance the raw data's quality, we conducted a two-step pre-processing stage. The first step involved linear interpolation, which not only handled irregular sampling but also addressed missing data points. This step was particularly important for the parameter estimation process. We set the new sampling rate to 30 seconds, striking a balance between performance evaluation and computational complexity. Choosing the optimal sampling rate was important, as smaller values led to slow optimization processes without significantly improving accuracy, while larger values compromised accuracy but accelerated optimization. In the second step of pre-processing, we aimed to handle anomalies caused by system errors or external factors. This was accomplished through the implementation of a moving median filter, effectively replacing anomalous data points with more representative values, thereby enhancing the accuracy of parameter estimation. The processed measurements for temperatures and mass flow rates for HE-2 are shown in Figure 3.

Two datasets were created from this processed data for gray-box model development and assessment, each comprising approximately 139 hours of data. The first dataset, collected from HE-2, was split into train and test sets. To investigate the impact of training/testing data splits on parameter estimation accuracy, we split the train and test sets for the same overall dataset into three different subgroupings (D1, D2, D3) as indicated in Table 1. This allowed us to investigate the quantity of training data required for accurate parameter estimation for HE-2. The second dataset (from HE-1) was used for validation and to evaluate the generalizability of our approach. It is worth noting that unlike the conventional approach in machine learning, where the train, test, and validation sets are sampled from the same dataset, we employed a different strategy here. The validation set was taken from another equipment, HE-1, even though theoretically, both HE-1 and HE-2 should be identical. The rationale behind this approach was to check the robustness and generalization of the identified grey-box models, considering the real-world scenario where perfect equipment matches may not always be feasible.

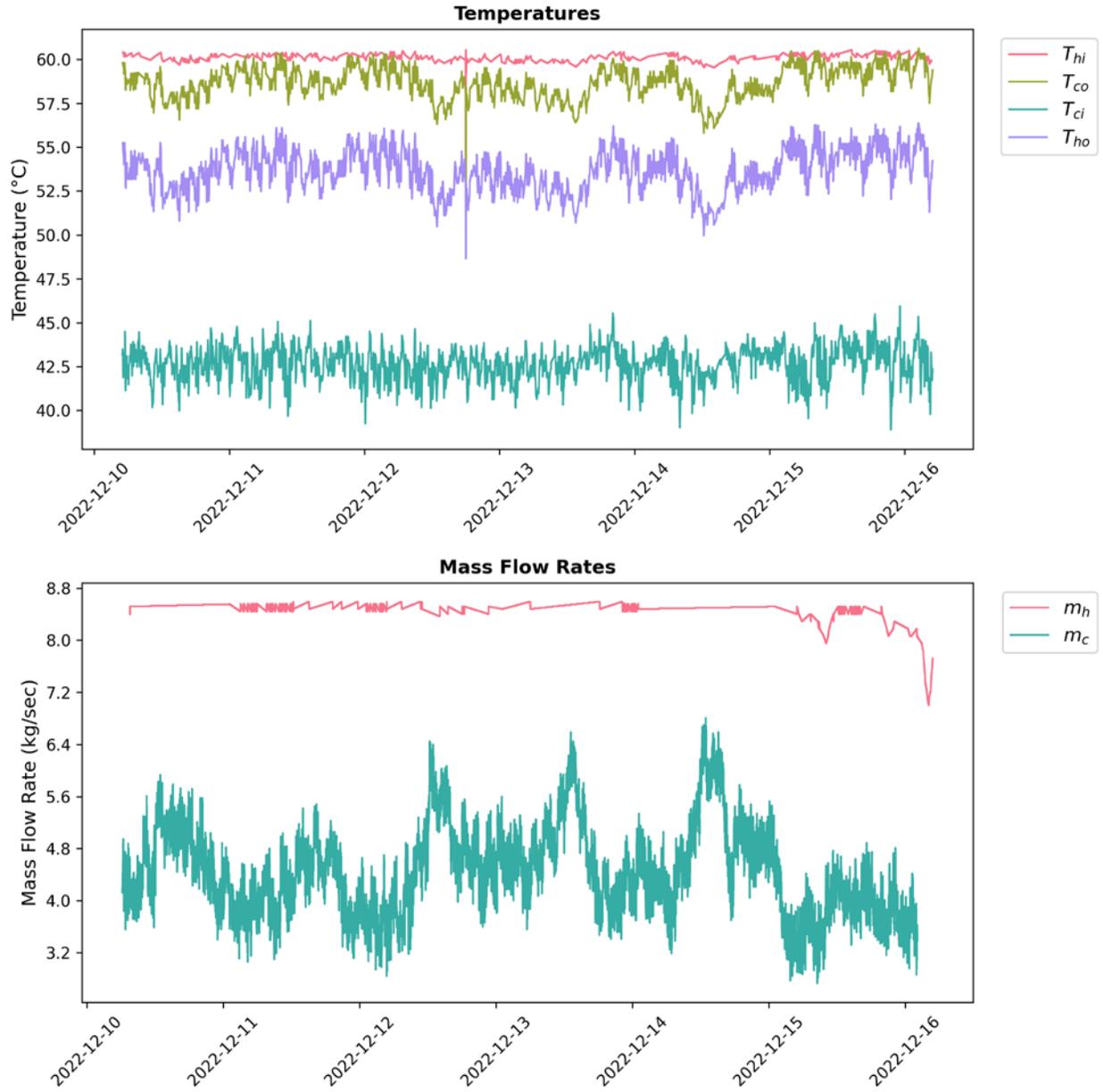

*Figure 3 Processed HE-2 temperatures (top) and mass flow rate (bottom) data*

*Table 1  Train and Test Splits Explored in this Study*

| Train split (hours) | Test split (hours) | Label |
|---|---|---|
| 11 | 128 | D1 |
| 25 | 114 | D2 |
| 53 | 86 | D3 |

## 3.4 Parameter Estimation

During the project's development and parameter estimation phase, we utilized a robust set of Python scientific computing packages for efficiency and reliability. Core packages included NumPy [48] for numerical computing, SciPy [49] for specialized scientific computations, Pandas [50] for data manipulation and pre-processing, and Matplotlib [51] for data visualization. These packages served as the foundation for data processing and model optimization. Central to our model formulation was the Pyomo package [52], which facilitated the construction of the Ordinary Differential Equation (ODE) model and offered a unified interface to various optimization solvers, including IPOPT.

The parameter estimation for the gray-box model, denoted by equations (3) and (4), was recast as an optimization problem with the objective of minimizing the mismatch between observed and simulated data. Utilizing the least squares method, a widely used technique in model optimization, we formulated an objective function ($J$) to minimize the sum of squared errors. The goal was to identify the optimal set of parameters, represented by the vector $k$, that offered the best fit to the observed data. The objective function ($J$) is formulated as follows:

$$J(k) = \sum (x - \hat{x}(k))^2 \qquad (5)$$

where $J(k)$ represents the objective function with respect to the parameter set. Here, $x$ denotes the vector representing the measured data, and $\hat{x}(k)$ is the vector representing the simulated model's output with parameter set $k$, signifying the states of the heat exchanger, see Figure 2. The parameter estimation workflow is summarized in Figure 4.

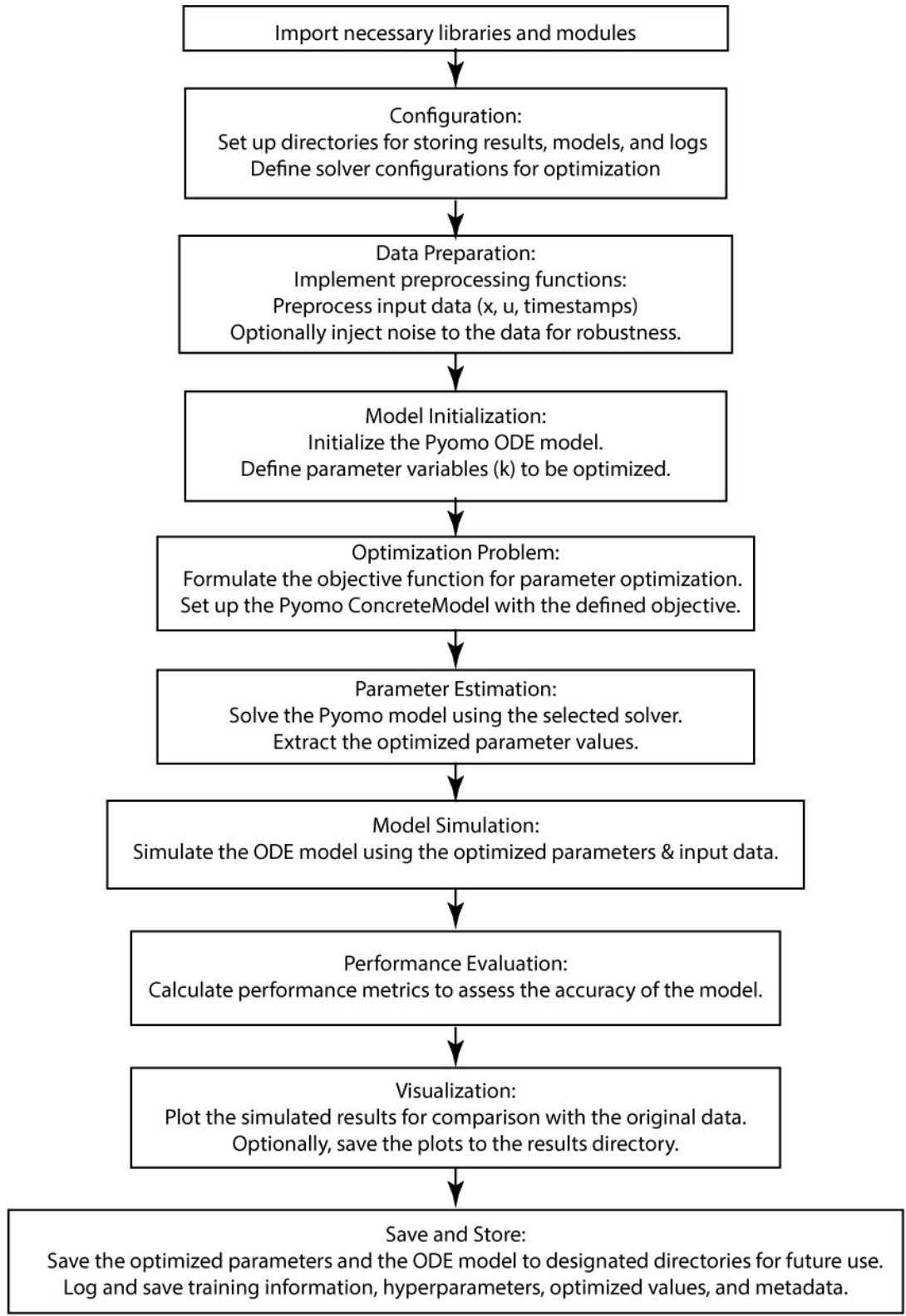

*Figure 4 Parameter Estimation Workflow (high-level representation of pseudocode)*

For routine data processing, simulations, and less computationally intensive tasks, a laptop equipped with a 6-core Intel Core i7 processor and 16 GB of device memory, running Windows 11 operating system was utilized. However, for computationally demanding tasks such as optimization and parameter estimation, the NEOS server was used, ensuring efficient use of computational power to tackle complex optimization challenges. During the simulation phase, the SciPy *odeint* function was employed to simulate the model's behavior. However, when it came to the optimization process, this was replaced by a simple Forward Euler integration method to improve computational efficiency during the optimization, ensuring a quicker convergence to optimal parameter values. Two models were developed in this study – a *vanilla* model without noise injection where the $x$ vector outputs were predicted and a noise-injected model, created as described in the following section.

## 3.5 Noise Injection

In scenarios where a model, such as our gray-box model, struggles to capture the full complexity of the data, noise injection can be utilized as a strategy to increase the model's performance. By introducing carefully calibrated random variations to our training data, we aim to nudge our gray-box model out of any local minima it may be stuck in during the training process. This is expected to improve the model's ability to learn and generalize the inherent dynamics of the system. A normal (Gaussian) distribution was used to generate noise for this study since it represents natural process variability well. Denoting the original output from the model as vector $x$, the noise-inject version $x_n$ can be defined as follows:

$$x_n = x + \mathcal{N}(0, \sigma^2) \qquad (6)$$

where $\mathcal{N}(0, \sigma^2)$ denotes a random Gaussian noise with mean = 0, variance $\sigma^2$, and a standard deviation $\sigma$, which is also known as the noise scale. Thus, $x_n$ is the original data with additive

Gaussian noise. This was injected into the output variables as shown in Figure 5, which illustrates that the noise-injected gray-box model will be trained to predict $x_n$ instead of $x$.

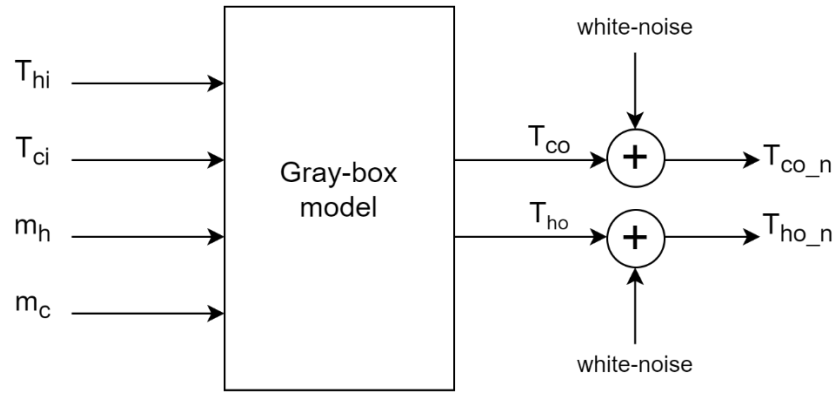

*Figure 5 Application of noise-injection to parameter estimation problem*

Note that while this technique introduces randomness to the model, its implementation was not random. Rather, the amount and type of noise added to the data was finely tuned. Too much noise can lead the model to fit the noise rather than the underlying pattern, a phenomenon known as overfitting, which would degrade the performance of the model. Conversely, too little noise might not provide sufficient impetus for the model to escape local minima. A range of noise scales was thus tested to identify the ideal value. Figure 6 compares a sample of as-measured data with the noise-injected data for three noise scales to show their relative impacts.

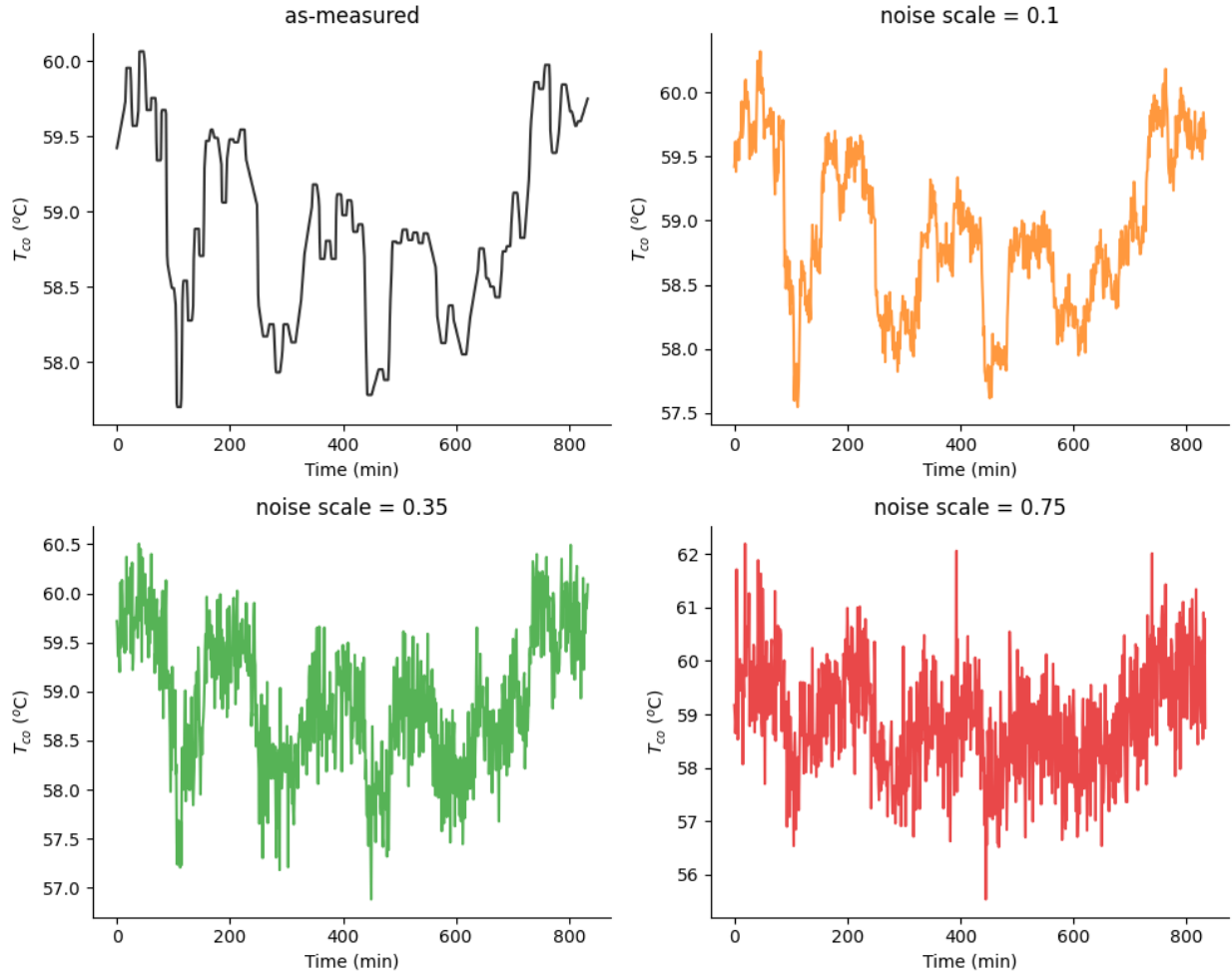

*Figure 6 Noise Injection Impact by Noise Scale (top left to bottom right: as-measured, noise scale = 0.1, noise scale = 0.35, noise scale = 0.75)*

### 3.6 Performance Evaluation

To evaluate model performance and compare results between the *vanilla* and noise-injected models, four metrics were utilized: Root Mean Squared Error (RMSE), Mean Absolute Percentage Error (MAPE), Maximum Absolute Error (Max_AE), and the coefficient of determination ($R^2$) coefficient. By using multiple metrics, we were able to compare the results from a variety of perspectives, gaining insight on the optimal degree of noise injection for gray-box model training.

# 4 Results

In this section, we present the results obtained from this study provide a comprehensive perspective on the role and effectiveness of noise injection in enhancing the predictive capabilities of dynamic gray-box modeling. This discussion will be structured as-follows: Section 4.1 presents the simulation results for the vanilla model; Section 4.2 presents the simulation results for noise-injected model; Section 4.3 discusses the tuning of noise-injection; and Section 4.4 summarizes the evaluated performance and comparative analysis.

## 4.1 Simulation Results for Vanilla Gray-Box Model

In our previous discussion, we categorized the collected data based on the splitting ratio between the train and test sets. In this subsection and the following one, we will focus on presenting simulation results derived from the model trained exclusively on group D2. Later on, we will utilize all three groups for a comprehensive comparative study. Figure 7 illustrates the simulation results obtained from the vanilla model trained on the D2 group data. It becomes apparent that the model lacks the capability to capture the system's dynamics and can only predict average values.

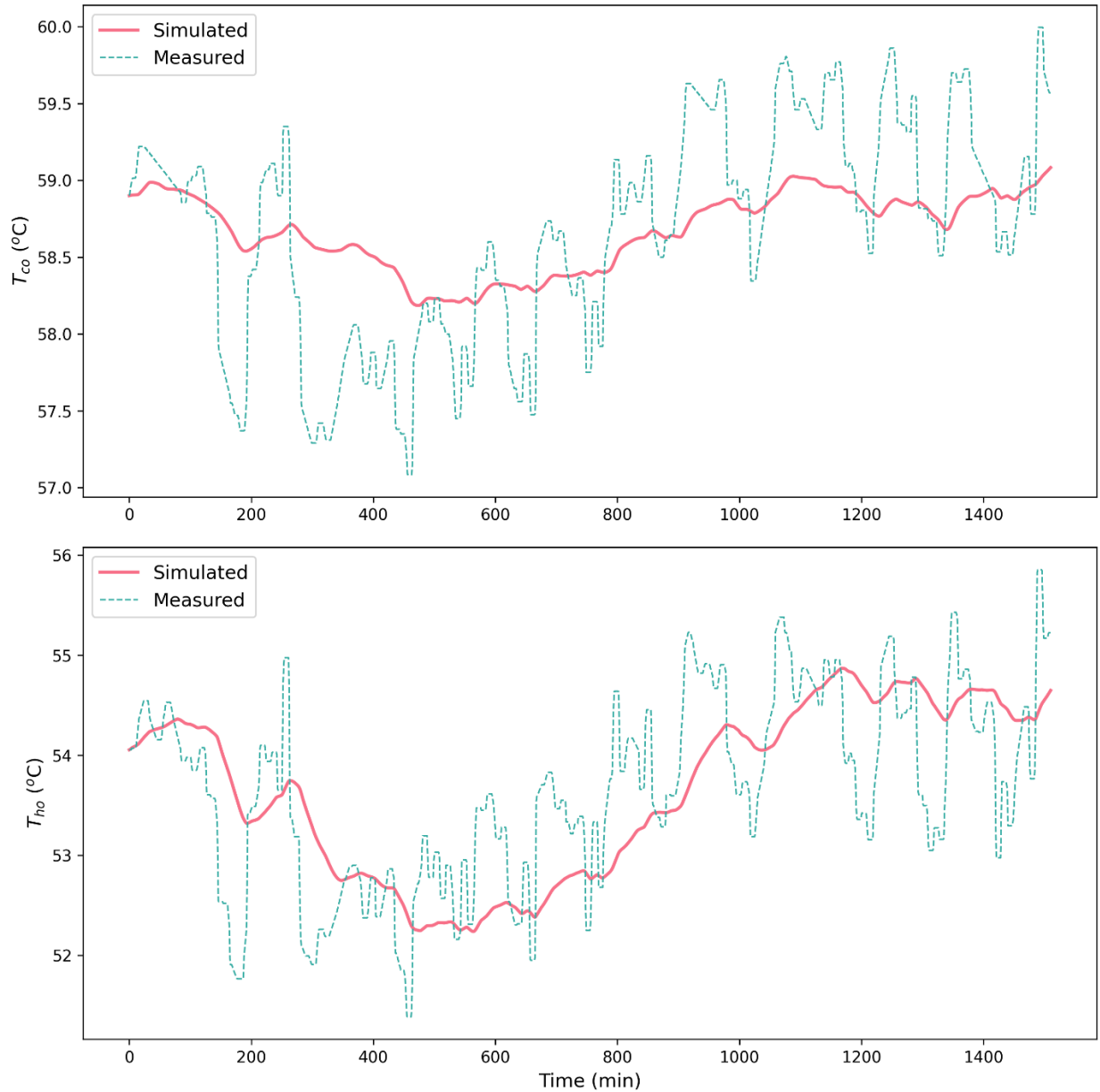

*Figure 7 Simulation response vs train data using vanilla model trained on D2 data*

To provide a complete analysis and facilitate a comparison with the proposed model, we also evaluated the performance of the vanilla model on the test and validation sets of group D2, as shown in Figure 8 and Figure 9, respectively. As anticipated, the model demonstrates poor performance on both of these sets. The results show that vanilla model could not fully replicate all the dynamics inherent in the training data. Instead, it was primarily able to provide an approximate representation of the average value.

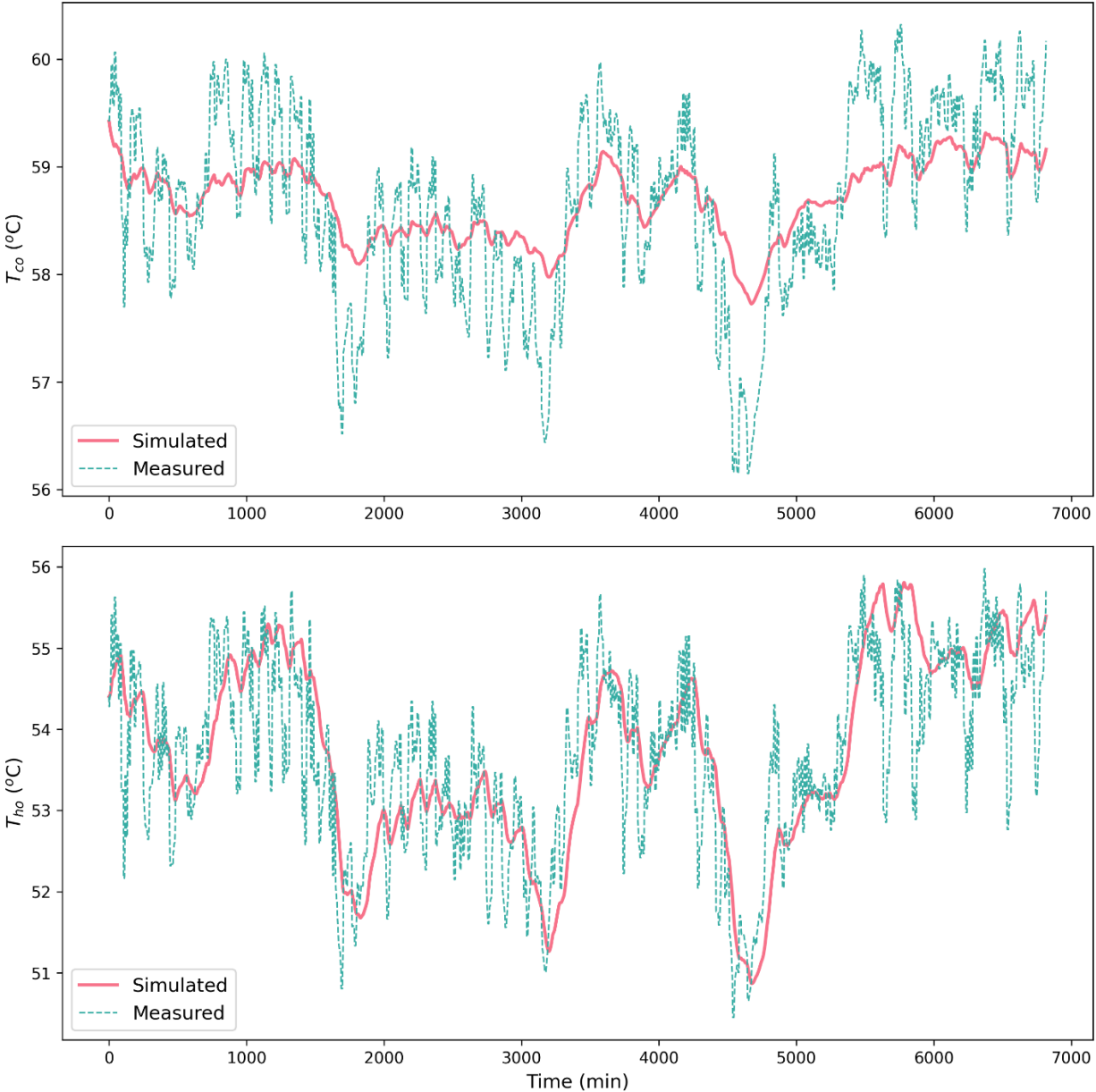

*Figure 8 Simulation response vs test data using vanilla model trained on D2 data*

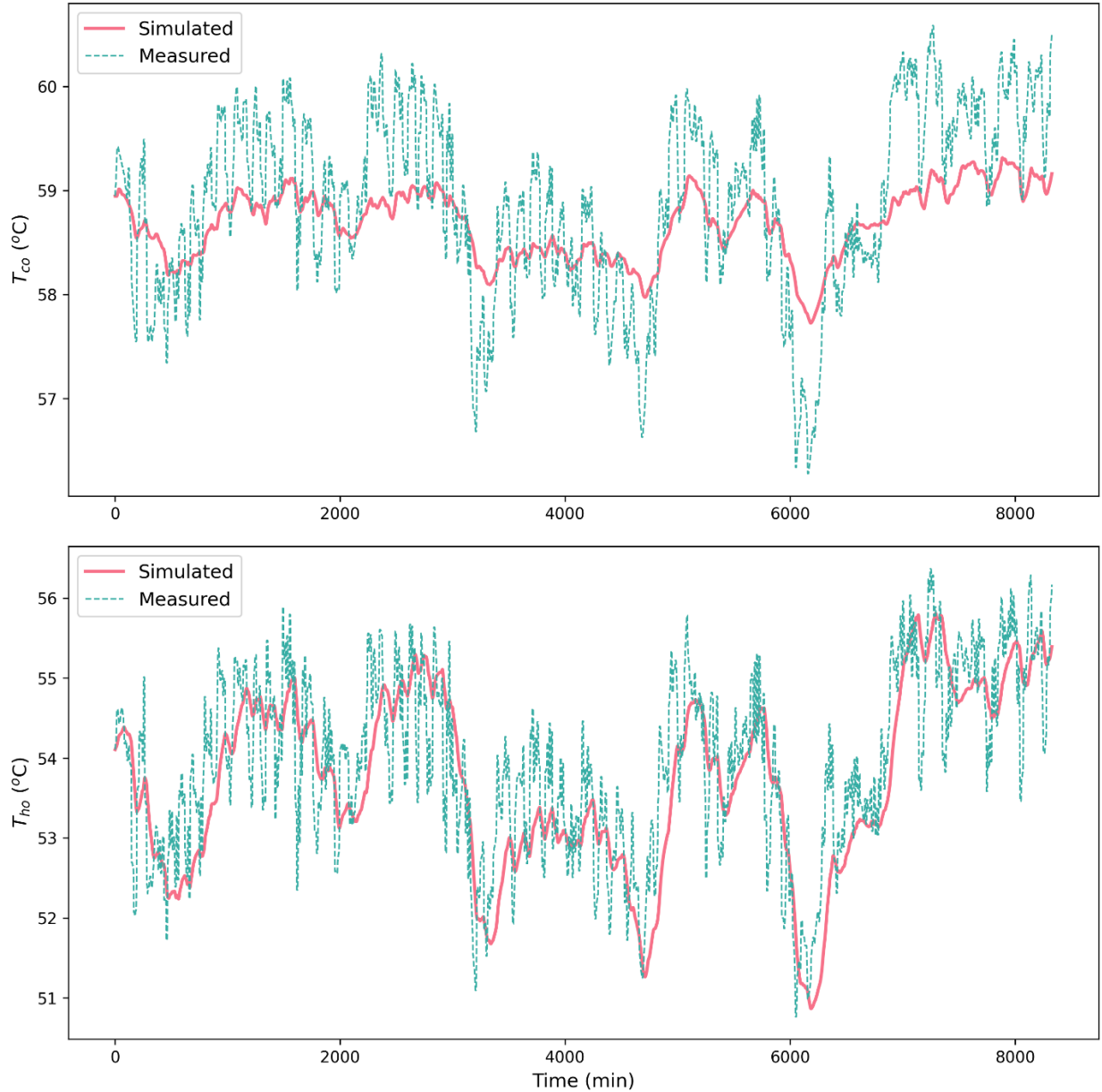

*Figure 9 Simulation response vs validation data using vanilla model trained on D2 data*

## 4.2 Simulation results for Noise-Injected Model

Just as for the vanilla model, we trained and evaluated the noise-injected model on group D2 of the datasets with noise-scale equal to 0.35, a choice substantiated in the next subsection. Figure 10 presents the simulation response compared to the actual measurements from the training dataset, showcasing the improved performance resulting from enhancing the model with noise injection. The noise-injected model exhibits a higher capability to capture system dynamics compared to the vanilla model.

Furthermore, Figure 10 and Figure 11 display the evaluation performance on the test and validation sets of group D2, respectively. Once again, through visual inspection, we observe a superior performance of the noise-injected model when compared to the vanilla model. The optimal estimates for the noise-injected model trained on D2 is given in Table 2.

*Table 2 Initial and optimal values for model parameters*

| Initial values | Optimal values |
| --- | --- |
| [0.1, 0.1, 0.1, 0.1] | [0.0284, 0.2218, 2.14, -1.1161] |

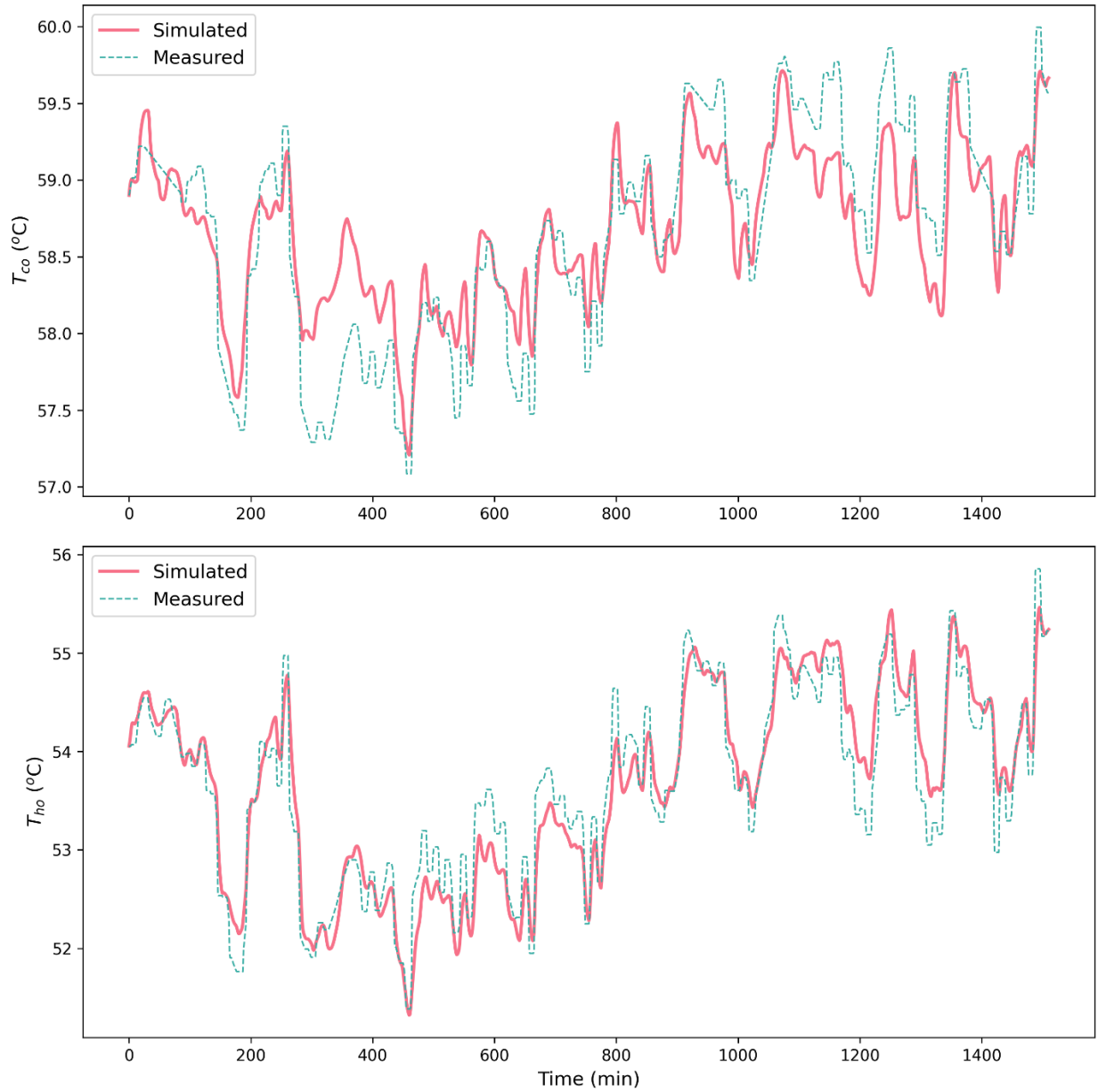

*Figure 10  Simulation response vs train data using noise-injected model trained on D2 data*

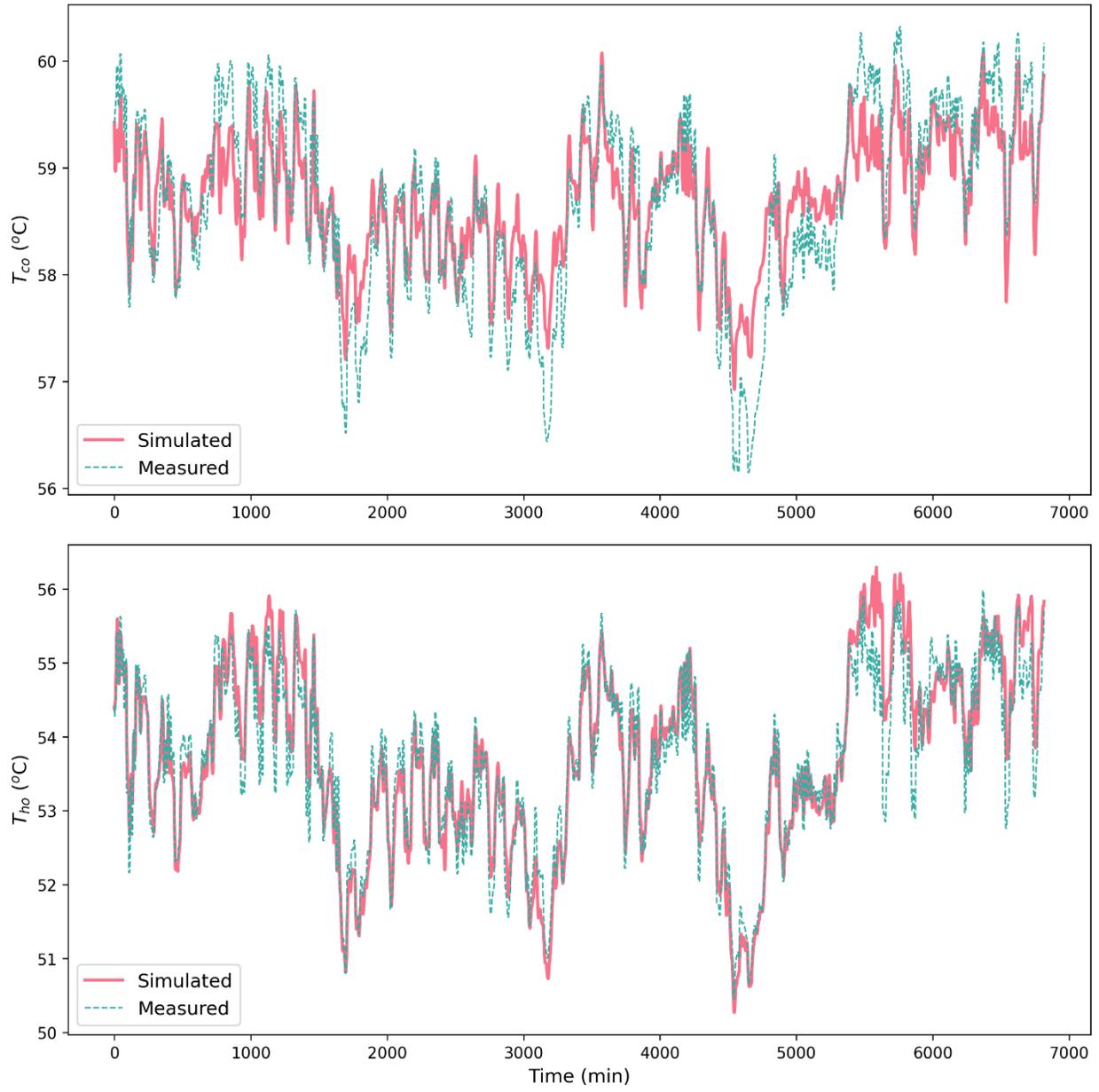

*Figure 11 Simulation response vs test data using noise-injected model trained on D2 data*

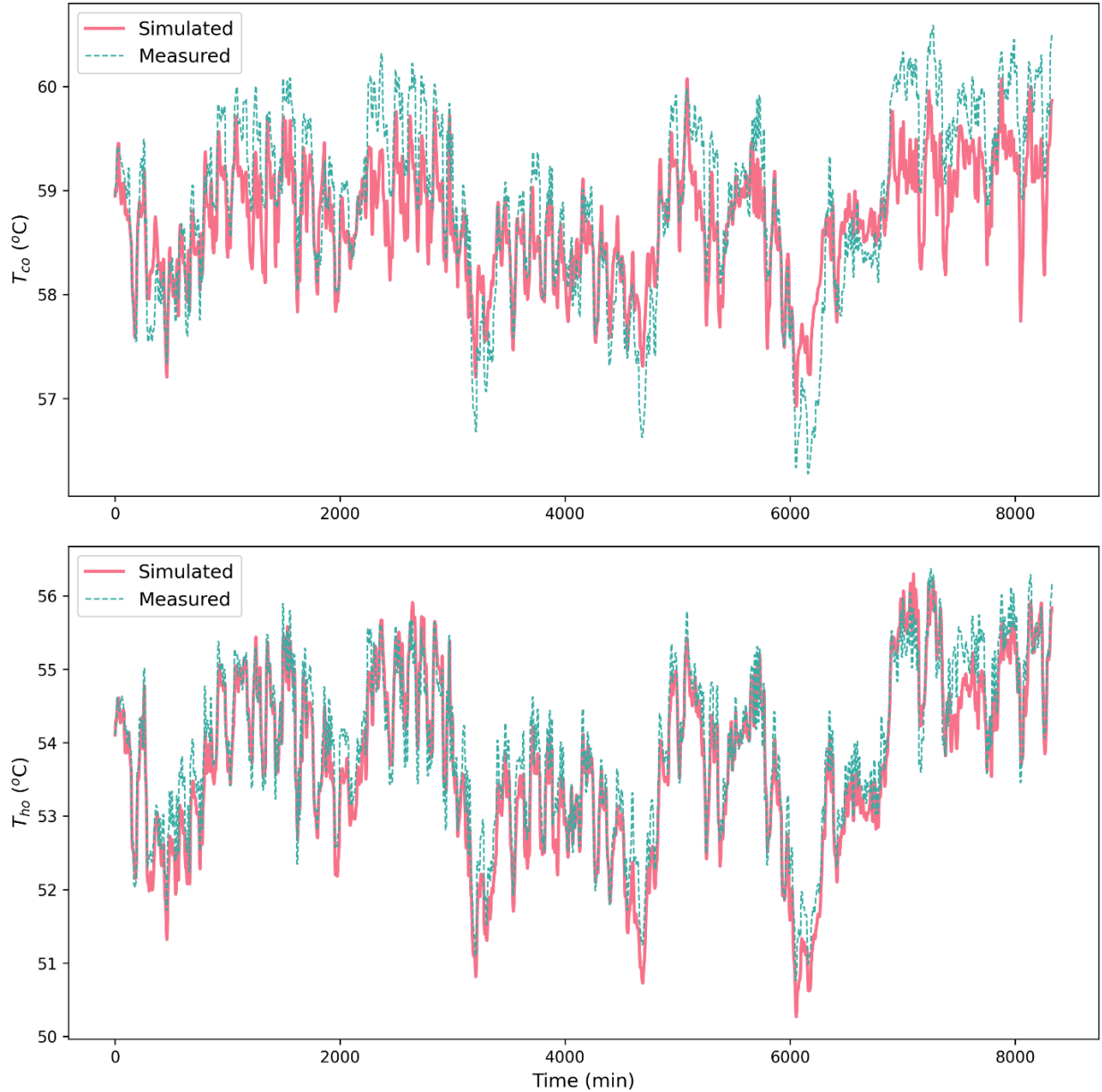

*Figure 12 Simulation response vs validation data using noise-injected model trained on D2 data*

## 4.3 Tuning Injected Noise

The effect of noise injection was investigated across a spectrum of noise scale values, ranging from 0.05 to 2.5. Figures 13 to 16 illustrate the performance various noise scales for each metric. To effectively showcase performance relative to the noise scale, the x-axis was constrained with a defined upper limit of 1.5 to ensure clear and concise presentation of the results. Upon visually analyzing these figures, it becomes evident that performance exhibits a substantial enhancement with an increase in the noise scale. However, it reaches a point of saturation, displaying a gradual decline thereafter. After careful consideration, the optimal value of 0.35 is chosen as the point that offers the best balance in performance.

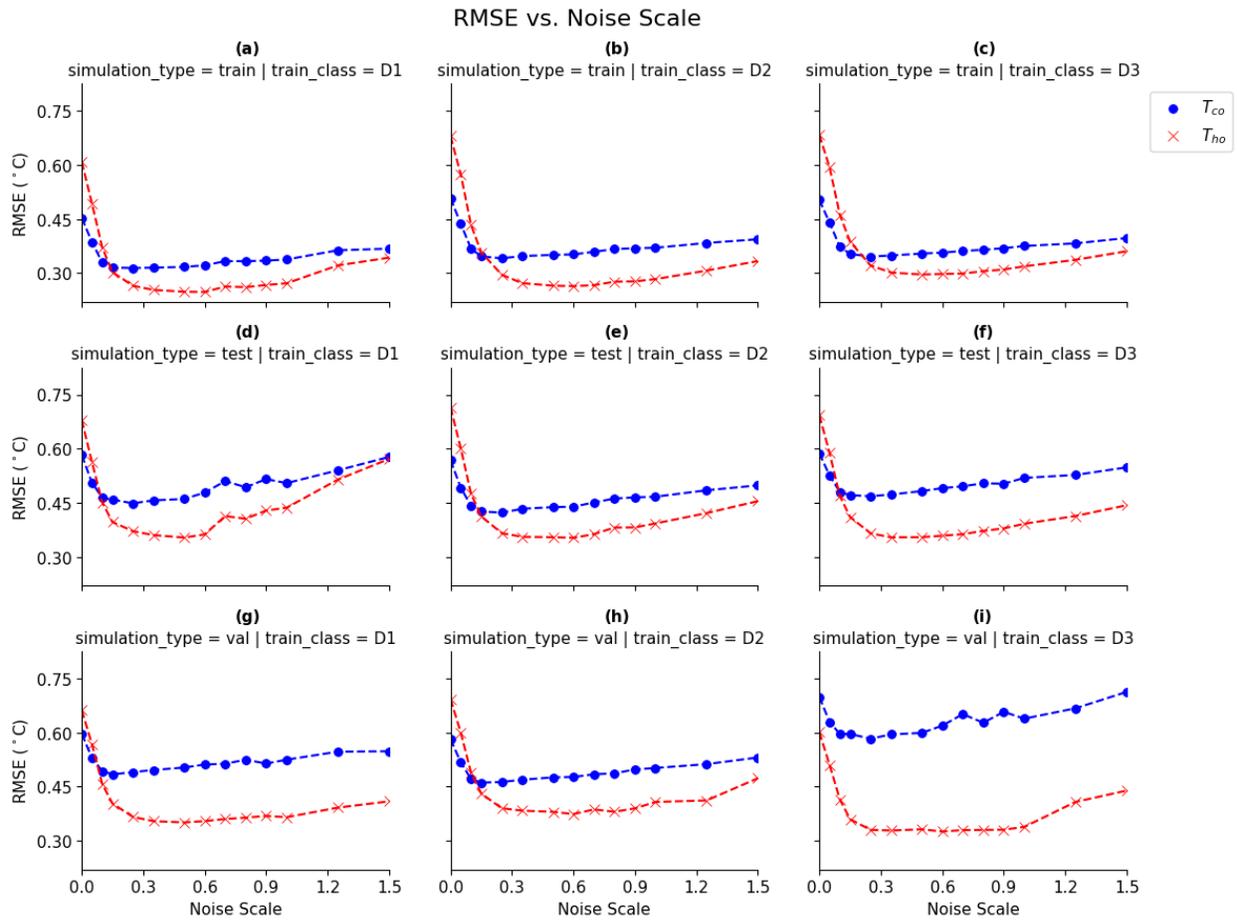

*Figure 13 RMSE vs Noise Scale (Note that a 0.00 value on the Noise Scale indicates the 'Vanilla' model)*

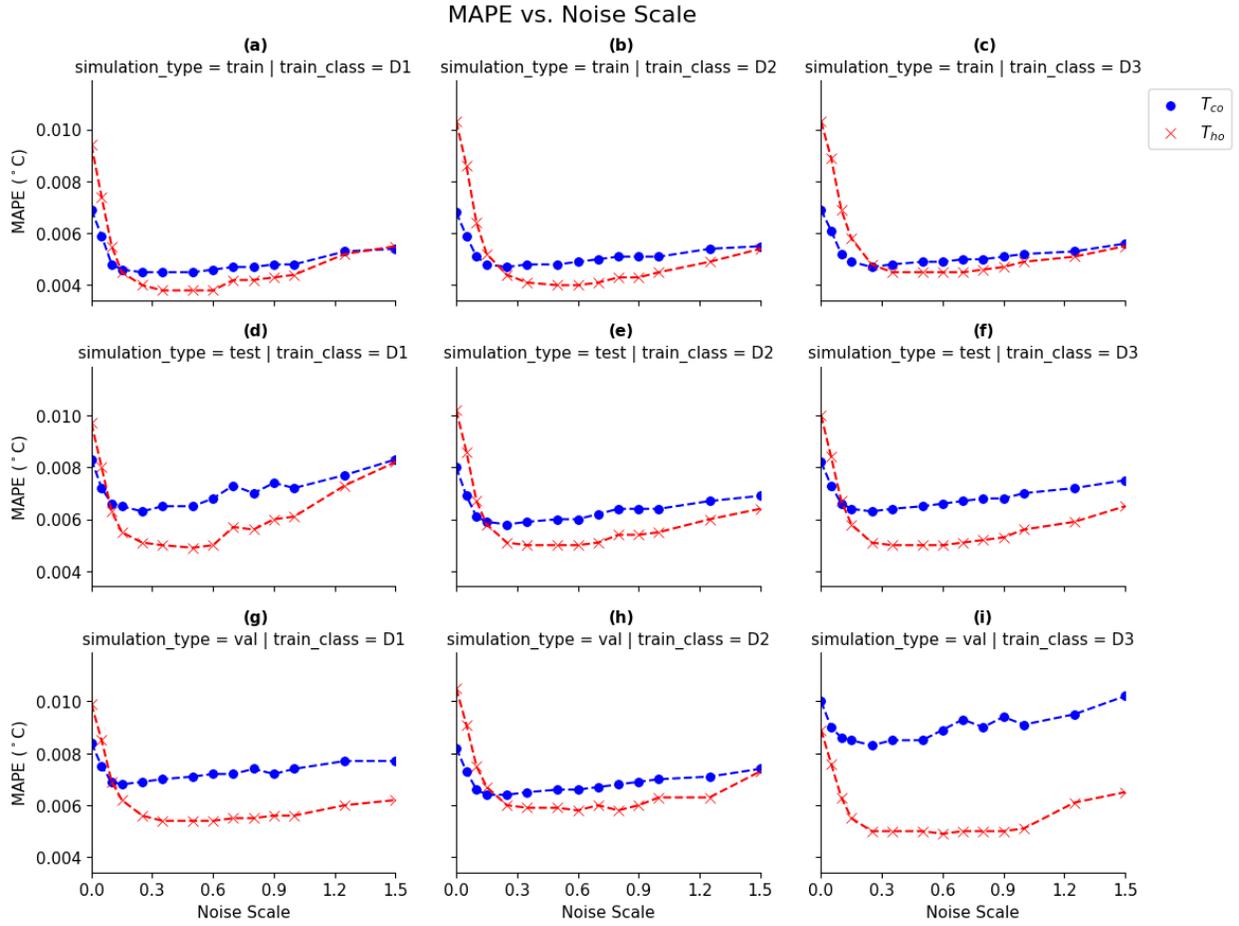

*Figure 14 MAPE vs Noise Scale*

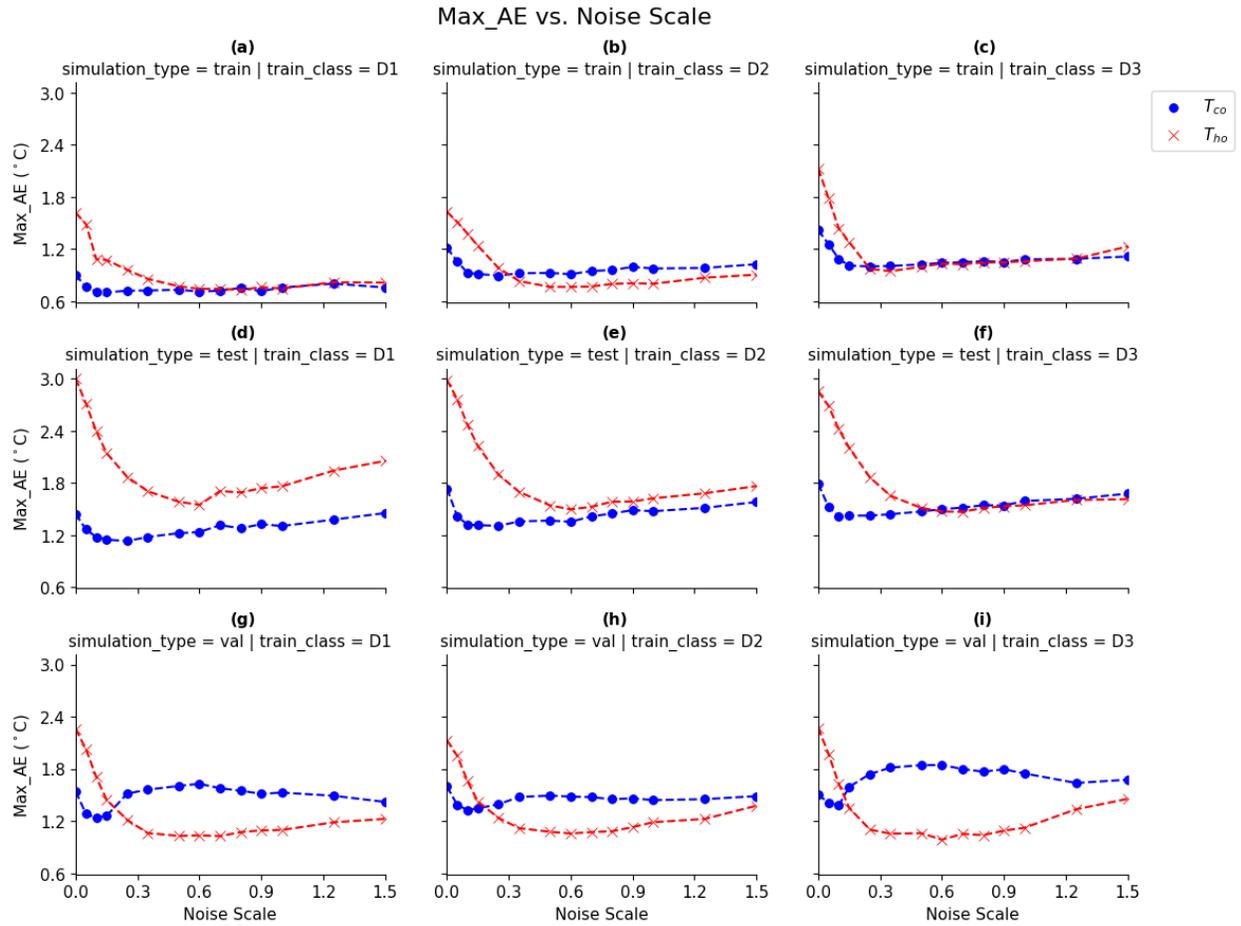

*Figure 15 Max_AE vs Noise Scale*

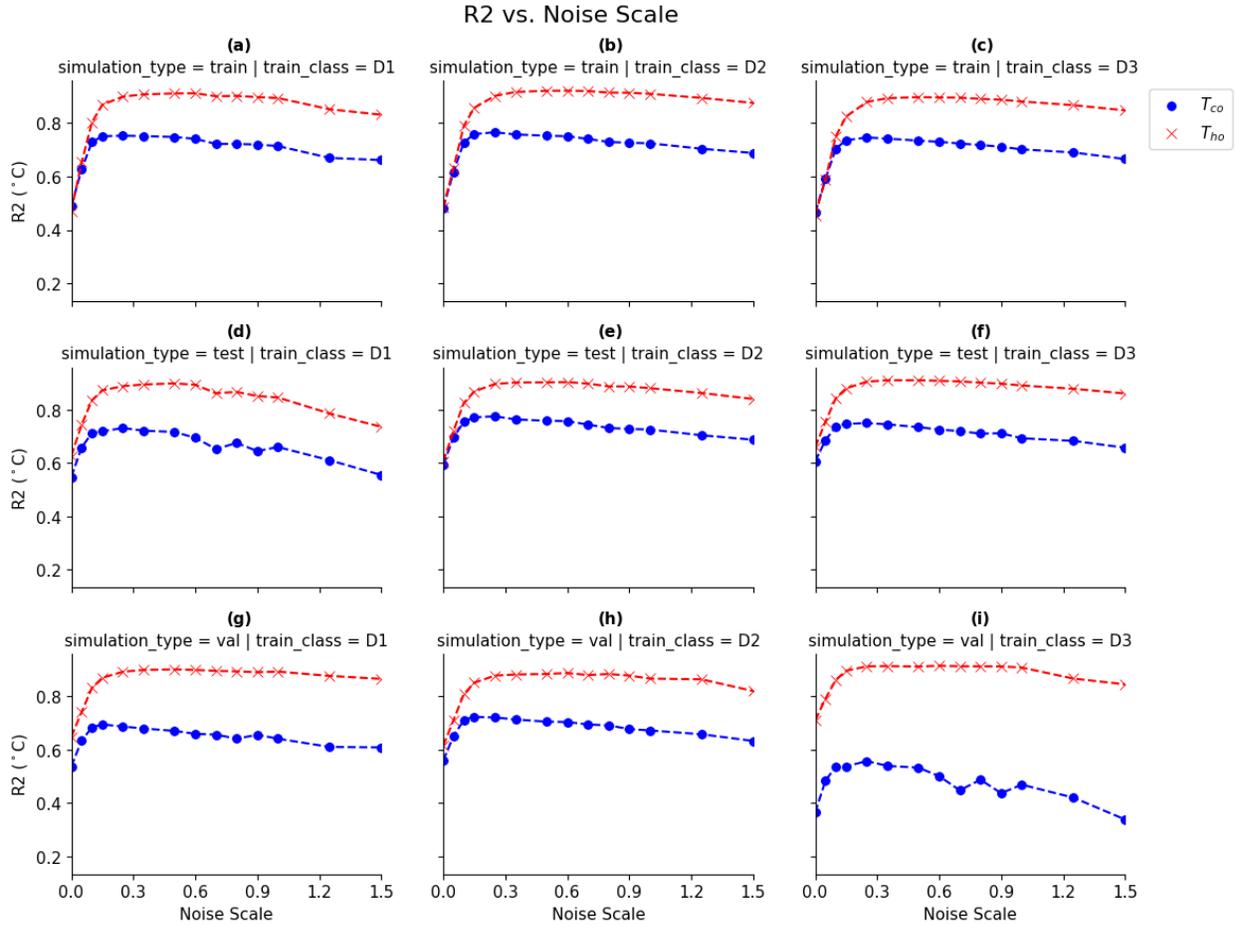

*Figure 16  $R^2$ vs Noise Scale*

**4.4   Performance evaluation and comparative analysis**

Tables 3, 4 and 5 summarize the best-performing noise injection values for each data set with the evaluation metrics for the supply side output temperature ($T_{co}$) and return side output ($T_{ho}$) temperature prediction for datasets D1, D2, and D3 respectively.

As shown in these tables, utilizing the noise injection optimal value (0.35) across datasets D1, D2, D3, remarkable improvements were observed. In training, for D1, RMSE and MAPE showed 58.36% and 59.57% improvements in $T_{ho}$, and 30.38% and 34.78% in $T_{co}$, respectively. In D2, the RMSE exhibited 59.96% improvement in $T_{ho}$ and 36.61% in $T_{co}$. Moreover, $R^2$ in D3 escalated by

96.55% in $T_{ho}$. Notably, the testing phase exhibited consistent enhancements, such as D1's 43.25% Max_AE and 48.2% MAE improvement in $T_{ho}$, and validation phase with D2's 47.43% Max_AE improvement in $T_{ho}$, underscoring generalizability. However, exceptions such as a 20.37% decline in Max_AE in D3's validation were observed, though other metrics showed improvement. Note that another decline, albeit smaller, was noted for D1, whereas an improvement in MAX_AE was observed for D2 for the validation.

*Table 3 Summary of Performance with Noise Injection Optimal Value (0.35) for Data Set D1*

| Run | metrics | $T\_ro$ | | | $T\_so$ | | |
|---|---|---|---|---|---|---|---|
| | | Vanilla | Noise-Injection | % Improvement | Vanilla | Noise-Injection | % Improvement |
| training | Max_AE | 1.6178 | 0.8573 | 47.01 | 0.9054 | 0.7246 | 19.97 |
| | MAE | 0.4952 | 0.2035 | 58.91 | 0.4026 | 0.2607 | 35.25 |
| | MAPE | 0.0094 | 0.0038 | 59.57 | 0.0069 | 0.0045 | 34.78 |
| | MSE | 0.3704 | 0.0642 | 82.67 | 0.2049 | 0.0993 | 51.54 |
| | RMSE | 0.6086 | 0.2534 | 58.36 | 0.4526 | 0.3151 | 30.38 |
| | R2 | 0.4705 | 0.9082 | 93.03 | 0.4875 | 0.7517 | 54.19 |
| testing | Max_AE | 3.0013 | 1.7032 | 43.25 | 1.444 | 1.1789 | 18.36 |
| | MAE | 0.5176 | 0.2681 | 48.2 | 0.487 | 0.3805 | 21.87 |
| | MAPE | 0.0097 | 0.0050 | 48.45 | 0.0083 | 0.0065 | 21.69 |
| | MSE | 0.4618 | 0.1300 | 71.85 | 0.3392 | 0.2087 | 38.47 |
| | RMSE | 0.6796 | 0.3606 | 46.94 | 0.5824 | 0.4569 | 21.55 |
| | R2 | 0.6269 | 0.8949 | 42.75 | 0.5476 | 0.7217 | 31.79 |
| validation | Max_AE | 2.2612 | 1.0623 | 53.02 | 1.5405 | 1.5616 | *-1.37* |
| | MAE | 0.5339 | 0.2923 | 45.25 | 0.4982 | 0.4106 | 17.58 |
| | MAPE | 0.0099 | 0.0054 | 45.45 | 0.0084 | 0.007 | 16.67 |
| | MSE | 0.4387 | 0.1257 | 71.35 | 0.3556 | 0.2463 | 30.74 |
| | RMSE | 0.6623 | 0.3545 | 46.47 | 0.5963 | 0.4963 | 16.77 |
| | R2 | 0.6480 | 0.8991 | 38.75 | 0.537 | 0.6794 | 26.52 |

*Table 4 Summary of Performance with Noise Injection Optimal Value (0.35) for Data Set D2*

| Run | metrics | T_ro | | | T_so | | |
|---|---|---|---|---|---|---|---|
| | | Vanilla | Noise-Injection | % Improvement | Vanilla | Noise-Injection | % Improvement |
| training | Max_AE | 1.6295 | 0.8331 | 48.87 | 1.216 | 0.9231 | 24.09 |
| | MAE | 0.5529 | 0.22 | 60.21 | 0.3972 | 0.2799 | 29.53 |
| | MAPE | 0.0103 | 0.0041 | 60.19 | 0.0068 | 0.0048 | 29.41 |
| | MSE | 0.4619 | 0.074 | 83.98 | 0.2579 | 0.1206 | 53.24 |
| | RMSE | 0.6796 | 0.2721 | 59.96 | 0.5078 | 0.3473 | 31.61 |
| | R2 | 0.4844 | 0.9174 | 89.39 | 0.4819 | 0.7577 | 57.23 |
| testing | Max_AE | 2.9831 | 1.6965 | 43.13 | 1.7344 | 1.3575 | 21.73 |
| | MAE | 0.5484 | 0.2677 | 51.19 | 0.4678 | 0.3467 | 25.89 |
| | MAPE | 0.0102 | 0.005 | 50.98 | 0.008 | 0.0059 | 26.25 |
| | MSE | 0.5112 | 0.1266 | 75.23 | 0.3228 | 0.1879 | 41.79 |
| | RMSE | 0.715 | 0.3559 | 50.22 | 0.5682 | 0.4335 | 23.71 |
| | R2 | 0.6056 | 0.9023 | 48.99 | 0.5944 | 0.7639 | 28.52 |
| validation | Max_AE | 2.1278 | 1.1185 | 47.43 | 1.5959 | 1.4767 | 7.47 |
| | MAE | 0.5673 | 0.3197 | 43.65 | 0.484 | 0.3837 | 20.72 |
| | MAPE | 0.0105 | 0.0059 | 43.81 | 0.0082 | 0.0065 | 20.73 |
| | MSE | 0.4805 | 0.147 | 69.41 | 0.3384 | 0.2204 | 34.87 |
| | RMSE | 0.6932 | 0.3835 | 44.68 | 0.5817 | 0.4694 | 19.31 |
| | R2 | 0.6144 | 0.882 | 43.55 | 0.5595 | 0.7131 | 27.45 |

*Table 5 Summary of Performance with Noise Injection Optimal Value (0.35) for Data Set D3*

| Run | metrics | T_ro | | | T_so | | |
|---|---|---|---|---|---|---|---|
| | | Vanilla | Noise-Injection | % Improvement | Vanilla | Noise-Injection | % Improvement |
| training | Max_AE | 2.1277 | 0.9466 | 55.51 | 1.4178 | 1.0064 | 29.02 |
| | MAE | 0.5524 | 0.2428 | 56.05 | 0.4073 | 0.281 | 31.01 |
| | MAPE | 0.0103 | 0.0045 | 56.31 | 0.0069 | 0.0048 | 30.43 |
| | MSE | 0.4672 | 0.0909 | 80.54 | 0.2531 | 0.1218 | 51.88 |
| | RMSE | 0.6835 | 0.3016 | 55.87 | 0.5031 | 0.3489 | 30.65 |
| | R2 | 0.4548 | 0.8939 | 96.55 | 0.4641 | 0.7422 | 59.92 |
| testing | Max_AE | 2.8525 | 1.6521 | 42.08 | 1.7904 | 1.4398 | 19.58 |
| | MAE | 0.5361 | 0.2678 | 50.05 | 0.4778 | 0.3717 | 22.21 |
| | MAPE | 0.01 | 0.0050 | 50 | 0.0082 | 0.0064 | 21.95 |
| | MSE | 0.4801 | 0.1257 | 73.82 | 0.3454 | 0.2239 | 35.18 |
| | RMSE | 0.6929 | 0.3545 | 48.84 | 0.5877 | 0.4731 | 19.5 |
| | R2 | 0.6596 | 0.9109 | 38.1 | 0.6068 | 0.7452 | 22.81 |

| validation | Max_AE | 2.2736 | 1.0551 | 53.59 | 1.5073 | 1.8144 | *-20.37* |
|---|---|---|---|---|---|---|---|
| | MAE | 0.4805 | 0.2688 | 44.06 | 0.5901 | 0.5025 | 14.84 |
| | MAPE | 0.0089 | 0.0050 | 43.82 | 0.01 | 0.0085 | 15 |
| | MSE | 0.3617 | 0.1078 | 70.2 | 0.4883 | 0.354 | 27.5 |
| | RMSE | 0.6014 | 0.3284 | 45.39 | 0.6988 | 0.595 | 14.85 |
| | R2 | 0.7097 | 0.9135 | 28.72 | 0.3643 | 0.5391 | 47.98 |

## 5 Discussion

The current study has explored the potential benefits of deliberate noise-injection techniques in addressing challenges associated with parameter estimation for dynamic gray-box modeling. The results obtained from our research highlight the efficacy of this approach in enhancing model generalization, handling uncertainties, and improving the accuracy and reliability of parameter estimation. It's important to note that noise was exclusively injected into the training data. This noise injection aimed to enable the optimization algorithm to efficiently determine the most appropriate parameters for the gray-box model. Following the completion of the optimization phase, which encompassed the training process, these derived optimal parameters were then employed in executing simulations on the non-noise-injected training, testing, and validation datasets. During these simulations, the optimal parameters are exclusively utilized for both all model variants and no additional noise was introduced to any of the data subsets, including training, testing, or validation data.

The simulation results provide compelling evidence of the benefits of noise injection. In comparison to the vanilla model, which struggles to capture the system's dynamics and is limited to predicting average values, the noise-injected model displays a significantly improved performance. The noise-injected model demonstrates a higher capability to capture the intricate dynamics of the system, both in training and evaluation stages. Our study represents a notable contribution to the field by emphasizing the significance of noise injection in dynamic gray-box modeling. While previous research has often neglected the potential advantages of controlled randomness, our work demonstrates its capability to empower dynamic gray-box models to explore diverse regions of the parameter space. By allowing the model to navigate through various possible scenarios, we observed improved robustness in parameter estimation, which is especially crucial for real physical systems characterized by nonlinearities and unmodeled dynamics.

The positive impact of noise injection across the three datasets demonstrates the generalizability of this approach. As discussed in the methodology, the testing and validation datasets were derived from two different sets of equipment, HE-1 and HE-2, rather than from a singular source, as is the conventional practice in machine learning. This deviation was intentional, serving as a mechanism to test the robustness and adaptability of the developed gray-box models in scenarios where equipment may not be perfectly identical, reflecting real-world challenges. In testing, improvements of 38%-53% (average 45%) were seen in metrics like Max_AE, MAE, and $R^2$, reflecting the ability of the algorithm to predict information for an unseen piece of equipment. Similar trends were mirrored in the validation set with similar degrees of improvement, with the exception of a notable decrease in Max_AE, which is at present unexplained but the remaining metrics showed substantial improvement. This consistency in predictions for the second heat exchanger demonstrates the method's ability to generalize well without overfitting the training data. However, the exceptions to this performance improvement signal the need for nuanced tuning and adaptation based on the particularities of each dataset and illustrate the need to make adjustments to the generalized model when applying it to new equipment.

## 6 Conclusions

Gray-box modeling represents a valuable approach in the development of equipment emulators, particularly in integrating physics into models for added confidence. Despite its potential, challenges like model nonlinearity, unmodeled dynamics, and local minima have hindered its performance. This study introduced a technique of injecting noise into the training dataset to address these uncertainties and enhance the robustness of gray-box creation.

Through the application of noise injection to a dynamic model of a water-water heat exchanger, the study demonstrated significant improvements in modeling error, amounting to a 60% enhancement on the training set and 50% and 45% on the test and validation sets, respectively. This method enabled a more accurate and reliable parameter estimation in gray-box models, reflecting a robust response to uncertainties. The findings highlight the necessity and potential of improving gray-box modeling techniques, underlining their vital role in equipment emulation and online controls optimization. The determined optimal noise-injection value across different performance evaluation metrics exhibits the consistency and effectiveness of this innovative approach. While the proposed technique has demonstrated notable advantages over the vanilla approach, one of the key limitations lies in the selection of the optimal noise-scale value. Determining this value *a priori* remains a challenge, as it significantly influences the performance improvement achieved through noise injection. In our current study, we systematically explored a wide range of noise-scale values, ranging from 0.05 to 2.50. Through thorough experimentation and evaluation, we identified the optimal noise-scale value as 0.35, based on predefined evaluation metrics. However, it is important to acknowledge that the optimal value may vary depending on the specific characteristics of different systems and applications. More extensive research across diverse domains is necessary to comprehensively understand the selection of the optimal noise-scale value. Fine-tuning this parameter for various scenarios will provide insights into the generalizability of the proposed technique and contribute to its wider applicability.

While this study focused on a specific equipment type, the insights gained are more broadly valuable. This study clearly demonstrates the value of noise injection to boost the overall performance of dynamic gray-box models, indicating a new avenue of research for advancing parameter estimation techniques and enhances the reliability and effectiveness of gray-box models across various domains. Future work should further explore the optimal settings and scenarios for noise injection, widening the understanding and applicability of this approach in various systems and applications. As the need for precise and reliable modeling continues to grow, the contributions of this paper underline the significance of investing in the development and refinement of gray-box modeling techniques. This research thus leaves a foundational footprint, steering the course towards more efficient dynamic gray-box modeling.